\documentclass[final,5p,]{elsarticle}

\usepackage[american]{babel}
\usepackage{afterpage}
 \usepackage{graphicx}

\usepackage{amssymb}
\usepackage{amsmath}
 \usepackage{amsthm}

 \usepackage{lineno}
 \setpagewiselinenumbers
\usepackage{subfigure}
\usepackage[percent]{overpic}
\usepackage{xspace}
\usepackage{layout}

\usepackage{algcompatible}
\usepackage{float}
\newfloat{algorithm}{tbhp}{loa}
\floatname{algorithm}{ALGORITHM}

\usepackage{soul,color,cancel}
\setstcolor{red}
\newcommand{\changed}[1]{{\color{blue}#1}}

\biboptions{sort&compress}

\graphicspath{{images/}}

\providecommand{\abs}[1]{\left\lvert#1\right\rvert}

\providecommand{\brac}[1]{\left(#1\right)}

\providecommand{\Brac}[1]{\left[#1\right]}

\providecommand{\var}[1]{\left<#1\right>}
\DeclareMathOperator{\grad}{grad}
\DeclareMathOperator*{\argmin}{argmin}

\DeclareMathOperator{\dx}{d\mathit{x}}

\DeclareMathOperator{\identity}{1\!\!1}
\providecommand{\smallfrac}[2]{\textstyle{\frac{#1}{#2}}}

\providecommand{\notinclude}[1]{}

\newcommand{\etal}{et al.\@\xspace}
\newcommand{\ie}{i.\thinspace e.\@\xspace}  
\newcommand{\eg}{e.\thinspace g.\@\xspace}

\newcommand{\edose}{\ensuremath{\mathrm{e}^-}/\ensuremath{\mathrm{\text{\AA}}^2}}

\journal{Ultramicroscopy}

\begin{document}
\begin{figure*}[t]
``NOTICE: this is the author's version of a work that was accepted for publication in Ultramicroscopy. Changes resulting from the publishing process, such as peer review, editing, corrections, structural formatting, and other quality control mechanisms may not be reflected in this document. Changes may have been made to this work since it was submitted for publication. A definitive version was subsequently published in \textbf{Ultramicroscopy, Volume 138, March 2014, Pages 46--56, DOI 10.1016/j.ultramic.2013.11.007}''
\end{figure*}
\thispagestyle{empty}
\setcounter{page}{0}
\newpage

\begin{frontmatter}



\title{Optimized imaging using non-rigid registration}


\author[IMI]{Benjamin~Berkels\fnref{label1}}
\ead{berkels@aices.rwth-aachen.de}
\fntext[label1]{Present Address: Aachen Institute for Advanced Study in Computational Engineering Science (AICES), RWTH Aachen University, Schinkelstr. 2, 52062 Aachen, Germany}
\author[IMI,Math]{Peter~Binev}
\ead{binev@math.sc.edu}
\author[Nano]{Douglas~A.~Blom\corref{cor1}}
\ead{doug.blom@sc.edu}
\cortext[cor1]{Correspondence to: University of South Carolina,
  NanoCenter, 1212 Greene St., Columbia, SC 29208, USA\\Tel:+1-803-777-2886;\\Fax:+1-803-777-8908;}
\author[IMI,Aachen]{Wolfgang~Dahmen}
\ead{dahmen@igpm.rwth-aachen.de}
\author[IMI,Math]{Robert~C.~Sharpley}
\ead{rcsharpley@gmail.com}
\author[IMI,Nano,Chem]{Thomas~Vogt}
\ead{tvogt@mailbox.sc.edu}
\address[IMI]{Interdisciplinary Mathematics Institute, 1523 Greene
  Street, University of South Carolina, Columbia, SC 29208, USA}
\address[Math]{Department of Mathematics, 1523 Greene Street,
  University of South Carolina, Columbia, SC 29208, USA}
\address[Nano]{NanoCenter, 1212 Greene Street, University of South Carolina,
  Columbia, SC 29208, USA}
\address[Aachen]{Institut f\"ur Geometrie und Praktische Mathematik,
  RWTH Aachen, Templergraben 55, 52056 Aachen, Germany}
\address[Chem]{Department of Chemistry and Biochemistry, 631 Sumter Street,
  University of South Carolina, Columbia, SC 29208, USA}

\begin{abstract}

  The extraordinary improvements of modern imaging devices offer
  access to data with unprecedented information content. However,
  widely used image processing methodologies fall far short of
  exploiting the full breadth of information offered by numerous types
  of scanning probe, optical, and electron microscopies. In many
  applications, it is necessary to keep measurement intensities below
  a desired threshold.  We propose a methodology for extracting an
  increased level of information by processing a series of data sets
  suffering, in particular, from high degree of spatial uncertainty
  caused by complex multiscale motion during the acquisition
  process. An important role is played by a nonrigid pixel-wise
  registration method that can cope with low signal-to-noise
  ratios. This is accompanied by formulating objective quality
  measures which replace human intervention and visual inspection in
  the processing chain.  Scanning transmission electron microscopy of
  siliceous zeolite material exhibits the above-mentioned obstructions
  and therefore serves as orientation and a test of our procedures.

\end{abstract}


\end{frontmatter}


\section{Introduction}
``Learning from data'' has become an indispensable pillar of science
of ever increasing importance.  Data acquisition can be broadly broken
down into two categories, parallel acquisition and serial acquisition.
In the case of serial acquisition when the change of signal between
data points encodes the information of primary interest (\eg images acquired one pixel at a time) extraction of the information from the raw data is difficult to achieve.  Obstacles include low signal-to-noise ratios, possible changes of the observed object due to the observation process, and - perhaps commonly less acknowledged - uncertainties in the positioning of the observations. As a consequence, the various stages of data processing  still often involve  a high degree of subjective human intervention.  This is particularly true, because standard methods cannot handle highly complex multiscale motion of the observed object. A representative example, providing the main orientation for this work, is {\em scanning transmission electron microscopy} (STEM). We are convinced though that the proposed methodology is {\em scale-invariant} and therefore relevant for a much wider scope of application.

The tremendous instrumental advances in STEM technology open new perspectives in understanding nanoscale properties of materials. However, in particular when dealing with beam sensitive materials, a full exploitation of this technology faces serious obstructions.  The acquisition of the images takes time during which both material damage as well as environmental disturbances can build up. A major consequence is a significant - relative to the scale under consideration -  highly complex motion intertwined with specimen distortion.

In response we propose a new data assimilation strategy that rests on the following two constituents:

\noindent
{\bf i)} replacing single frame high-dose data acquisition by taking multiple low-dose,  necessarily noisy, frames;

\noindent
{\bf ii)} properly synthesizing the information from such series of frames by a novel cascading registration methodology.

A few comments on the basic ingredients are needed.
Taking a single high-dose frame may in principle feature a higher
signal-to-noise ratio but typically increases the possibility of irreparable
  beam damage, prevents one from accurately tracking the combination of local and global motion during the scanning process, and possibly introduces additional unwanted physical artifacts.
  We use STEM imaging of a water-containing siliceous zeolite Y (Si-Y) as a representative guiding example, since it poses many of the obstacles which occur in practice when determining the structure and function of important classes of materials. In particular, Si-Y mimicks biological matter by rapidly deteriorating when exposed to high energy electron beams. Furthermore, the registration task for Si-Y poses difficult hurdles for registration which are often encountered in imaging materials, including a high degree of symmetry and low variety of structures in the presence of low signal-to-noise samples.

The core constituent~{\bf i)} above, i.e. taking
several low-dose short-exposure frames of measurements from
the same physical locations of the
specimen, allows us instead to better resolve the complex motion and nonstationary distortions.
 We demonstrate how the information from
multiple frames can be used to offset such distortions as those described.

 A critical ingredient of core constituent~{\bf ii)}  is to align the series of frames.
Extracting improved information from series of low quality images is a well-known concept, usually termed
{\em super resolution} in the imaging community. It rests in an essential way on the ability of accurately tracking inter-frame motion, typically using feature extraction.
However, the low signal-to-noise ratios prohibit accurate feature extraction.
Moreover, the physical nature of the motion encountered in series of STEM images is of a highly complex nature
that can therefore not be treated  satisfactorily by standard techniques.
The extreme sensitivity of STEM to the measuring environment gives rise to  motion which appears to exhibit at least three scales \cite{BinevNLM}: jitter  on the atomic level, contortion on the unit cell level, and a macro drift. As explained later in more detail, short time exposures provide the opportunity for a sufficiently refined temporal discretization. Furthermore,
being able to generate highly accurate pixel to pixel mappings, in contrast to  fiducial-based registration in
imaging, seems to be the only hope to avoid significant blurring at the image assimilation stage.

The few existing approaches in the STEM context resort to manual {\em
  rigid} alignment (see \eg in tomography \cite{TomographyMicro} and
single particle analysis \cite{Nickell2005227}).  Although in our
experiments, from a subjective point of view, rigid alignment appears
to yield good results, they are, however, as is typical, confined to a
relatively small image portion that is implicitly used as a feature
substitute.  First, this introduces a highly subjective and therefore
non-reproducible component into the data processing chain. Second, it
wastes and even obscures information about the observed object carried
by large parts of the image series. In particular, in STEM there is
often high interest in determining possible irregularities in the
atomic structure outside the small registration zone.

In order to be able to make full use of the information carried by the
frames and, in particular, to be able to reliably detect material
imperfections away from the subjective reference region we propose a
powerful {\em multilevel non-rigid registration} method accurately
linking the information carried by many sequentially acquired low-dose
frames. This gives rise to data assimilation without significant
blurring.  Furthermore, it serves an important universal objective,
namely to replace ``human weak links'' in the data processing chain by
more robust, highly accurate, quantifiable, and reproducible
processing modules.In this context a further important issue is to
formulate and apply quantitative objective quality measures to
illustrate the effectiveness of the procedures.

\section{Materials and Methods}
\subsection{Zeolites as Proxy Materials}
\label{sec:si:zeolites}
Zeolites are beam sensitive, crystalline aluminosilicates consisting
of AlO$_{4}$ and SiO$_{4}$ corner-sharing tetrahedra, whose
arrangement allows for a large variety of structures with different
pore sizes and pore connectivities. These materials are known to
deteriorate rapidly, due to a combination of radiolysis and knock-on
damage \cite{Ugurlu,Treacy,JMTNature}, when exposed to electron beams
generated by accelerating voltages between 60--200~kV.

 {\em Knock-on damage} occurs when the fast electrons impart sufficient energy to the atoms in the specimen to displace them from their equilibrium positions, thereby directly breaking the long-range order of the material.  {\em Radiolysis} occurs when the fast electrons ionize the atoms in the material which then subsequently lose their long-range order as they seek a lower energy state.
The ionization of water by the electron beam is thought to be a key step in the damage mechanism~\cite{Treacy,JMTNature}.
 Several excellent review articles on the application of transmission electron microscopy to the study of the structure of zeolites and other mesoporous materials have been recently published~\cite{Pan,JMTReview,DiazReview}.

Our choice of zeolites as proxies for imaging beam-sensitive materials is based on four properties:
(i)  high crystallographic symmetry and smaller unit cell dimensions which, compared to biological objects, simplify the quantification of results and allow us to focus on technique development, (ii) the presence of water, (iii) reduced structural variety and (iv) high beam sensitivity.
Each of these properties pose a significant challenge for the registration task.

\subsection{STEM Imaging}
\label{subsec:STEMImaging}
\begin{figure*}[ht]
\begin{center}
\begin{tabular}{cc}
\begin{overpic}[width=0.475\linewidth]{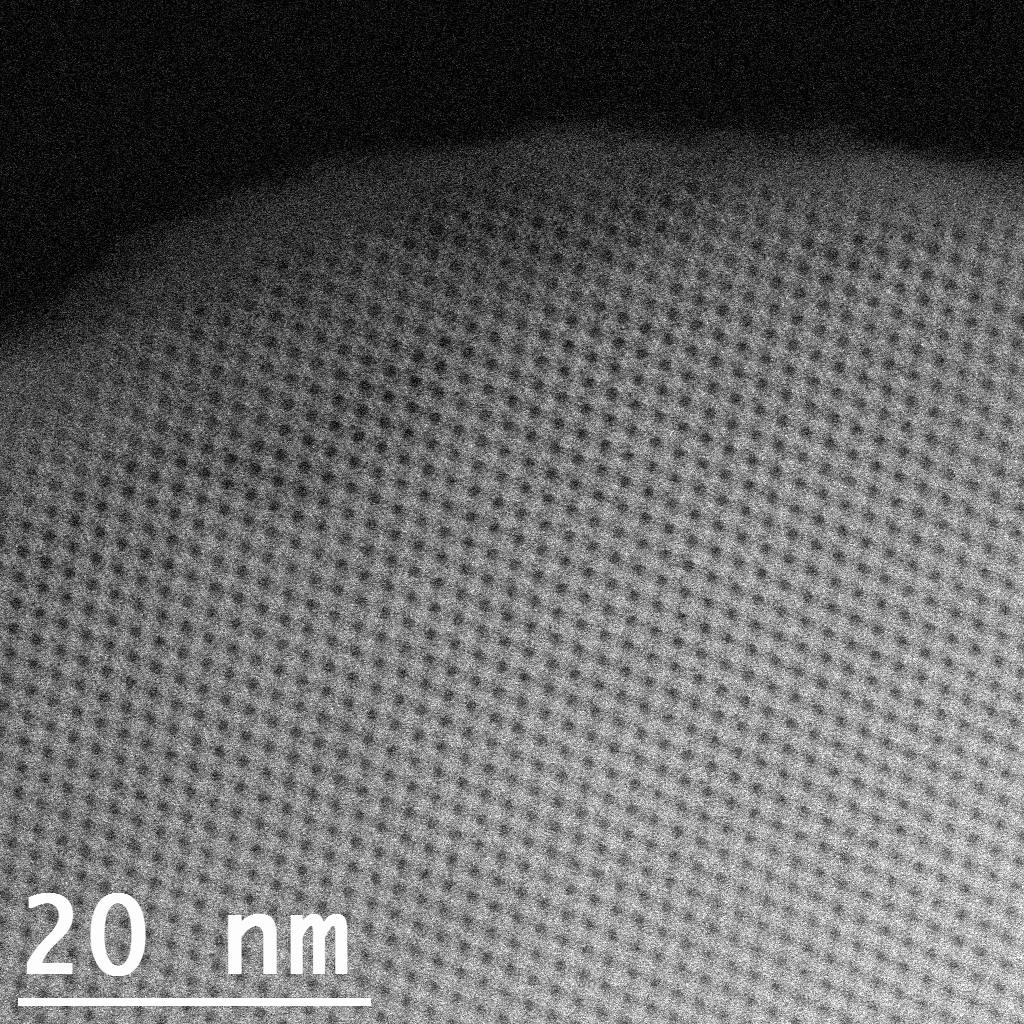}
\put(2,94){\colorbox{white}{\Large a}}\end{overpic}&
\begin{overpic}[width=0.475
\linewidth]{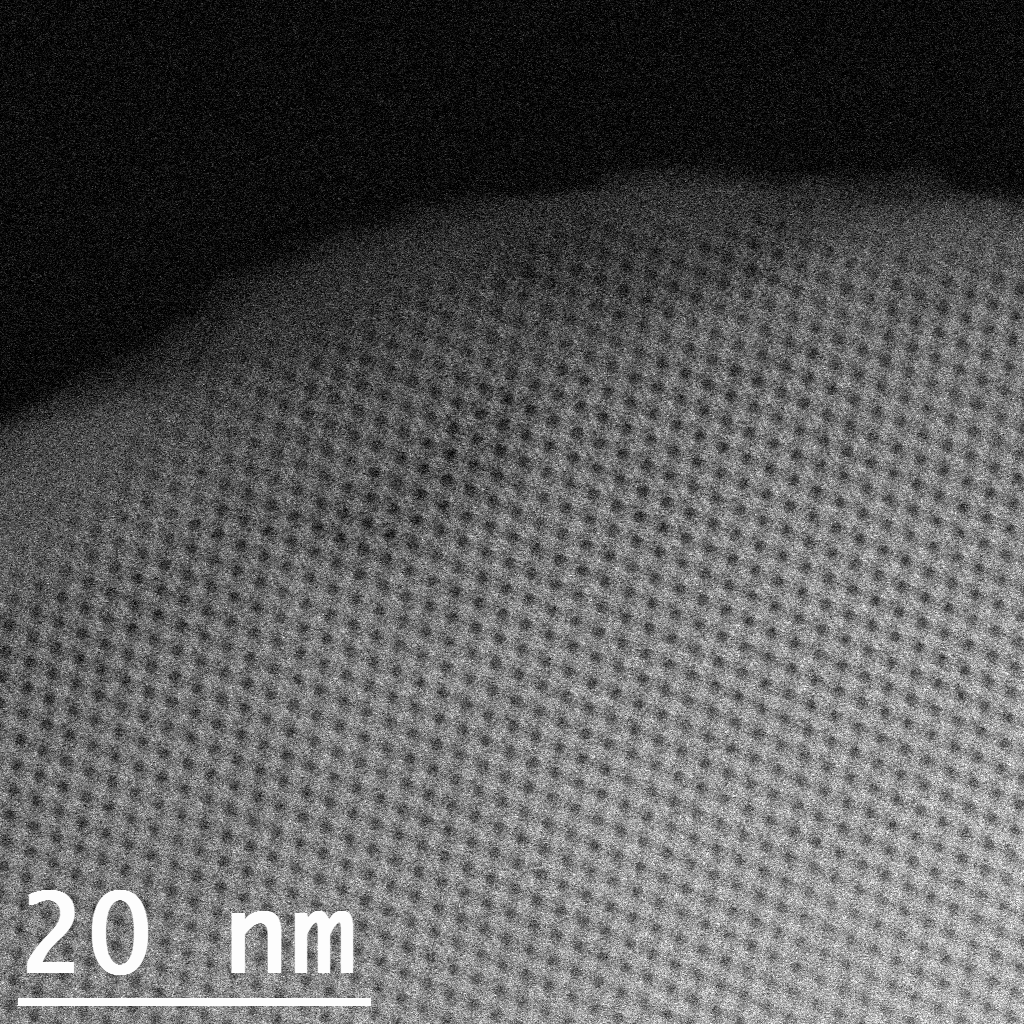}
\put(2,94){\colorbox{white}{\Large b}} \end{overpic}
\end{tabular}
\end{center}
\caption{First {\bf (a)} and ninth {\bf (b)} raw frames of a HAADF STEM zeolite Si-Y series, indicating significant sample drift as well as beam damage in the region near the material's thin edge.}
\label{fig:si:ZeoliteInputFrames}
\end{figure*}
STEM of zeolites has been used generally for either the high-contrast imaging of small metal nanoclusters within the pores using the high-angle annular dark-field (HAADF) technique or high spatial resolution chemical analysis using either energy dispersed \mbox{X-ray} (EDX) spectroscopy or electron energy loss spectroscopy (EELS)~\cite{JMTSTEM,JMTEELS}.
 Aberration-corrected STEM allows for the formation of sub-\AA ~probes with greatly increased brightness~\cite{Krivanek} which results in images with substantially better signal-to-noise ratio.  The aberration corrector nearly eliminates the electrons in the probe outside the central maximum which results in much more dose-efficient imaging.

 Ortalan \etal used an aberration-corrected STEM to image the location of single atoms of Iridium in a zeolite using a combination of low-dose image acquisition methodology and both Fourier filtering and real-space averaging~\cite{OrtalanHY}.  The  reported dose is similar to that in our work.

 For our experiments HAADF STEM images at 200~kV were recorded with a
 JEOL JEM 2100F TEM/STEM equip\-ped with a CEOS CESCOR aberration
 corrector. All axial aberrations of the electron wave were measured
 and corrected up to third order. Fifth order spherical aberration was
 minimized as well.  The illumination semi-angle was 15.5\,mrad, which
 at 200\,kV yields a nominal probe size of 0.1~nm.  The HAADF detector
 recorded electrons scattered between 50 and 284 mrad.  The probe
 current was 10\,pA as measured with a picoammeter attached to the
 small fluorescent focusing screen of the microscope.  The pixel size
 was 0.3~\AA$^{2}$ and dwell time per pixel was 7$\mu$s yielding an
 electron dose for the images of $\approx1400\,\edose$.  Image
 acquisition was controlled by a custom script in Digital Micrograph
 (Gatan, Pleasanton, CA) which sequentially recorded and saved a
 series of HAADF STEM images 1024$\times$1024 in size.

Our material test sample is a siliceous zeolite Y~(c.f. \cite{SI-Y})
which was completely de-aluminated. The Si-Y zeolite powder was
dispersed onto an amorphous carbon holey support film on a copper mesh
TEM grid.  A small particle was located and oriented along the
$<$110$>$ direction of the zeolite Y structure.
Figure~\ref{fig:si:ZeoliteInputFrames} shows the first and ninth
frames of a series of HAADF STEM images of the Si-Y zeolite sample
collected with the conditions described above.
Because of the relatively low electron dose the individual images in
Figure~\ref{fig:si:ZeoliteInputFrames} are quite noisy.  The largest
pores of the zeolite Y structure are evident, but details of the
structure are obscured by the noise level.  The overall drift of the
sample between the first and ninth frame is evident in the figure.
For the purpose of clarifying the atomic structure of the zeolite as
well as the noise level of the acquired HAADF STEM images we focus on
an area 5.1$\times$7.3~nm in size oriented such that $<$001$>$ is
vertical as shown in Figure~\ref{fig:si:multislice}\,a.  In the
acquired image, the super-cages of the zeolite are clear, but the
atomic structure of the framework is obscured due to the noise in the
image.

\subsection{Frozen Phonon Simulation of HAADF STEM image}
\label{sec:si:HAADFSTEMImageSimulation}

\begin{figure}[h]
\begin{center}
\begin{tabular}{cc}
\begin{overpic}[width=0.465\linewidth]{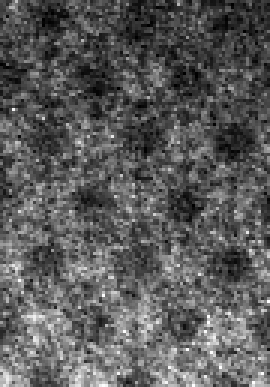} \put(2,94)
  {\colorbox{white}{\Large a}}\end{overpic}&
\begin{overpic}[width=0.465\linewidth]{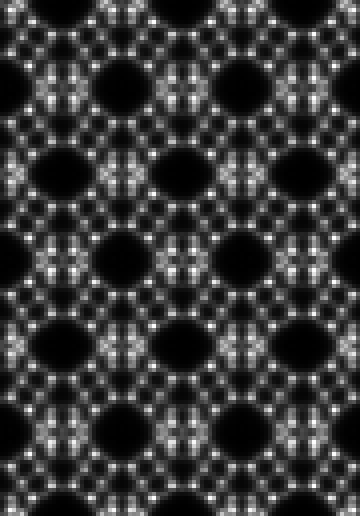}
  \put(2,94){\colorbox{white}{\Large b}}\end{overpic}
\end{tabular}
\end{center}
\caption{{\bf (a)} 5.1$\times$7.3~nm region of the first acquired
  low-dose frame rotated so that $<$001$>$ is vertical. {\bf (b)}
  Equivalent area and orientation of a HAADF STEM image simulation of
  the zeolite Si-Y in the $<$110$>$ zone axis orientation clearly
  revealing the large $\alpha$ cages and the double six-member rings.}
\label{fig:si:multislice}
\end{figure}

Figure~\ref{fig:si:multislice}\,b shows the results of a multislice
HAADF STEM image simulation using Kirkland's code~\cite{Kirkland}.
The simulation is equivalent to the input experimental images reported
here: 200\,kV, 15.5\,mrad convergence semi-angle, C$_{s}$ = 3\,$\mu$m,
C$_{5}$ = 0\,mm with a defocus value chosen to maximize the contrast
in the image (-20\,\AA~here).  The simulation was done with a
super-cell of \,68.4\,\AA$\times$72.8\,\AA~by 120\,\AA\, thick. A
total of 64~phonon configurations were needed to achieve convergence
of the calculation.  The slice thickness was set at slightly larger
than 1\,\AA. The HAADF detector spanned 50--284\,mrad.  Incoherent
blurring was added by convolving the simulation results with the
gaussian of 0.1~nm FWHM. Calculations were done for an area slightly
larger than 1/4 of the projected unit cell size of Si-Y $<$110$>$ and
then appropriate symmetry was applied to generate an image
5.1$\times$7.3~nm in size.  In this orientation the sodalite ($\beta$)
cages overlap and are not easily visualized.  The largest
$\alpha$-cages with a diameter of $\approx$\,13\,\AA\, are evident and
the $<$001$>$ direction is along the vertical direction in the Figure.
The double six-member rings run between the super-cages along
directions 60$^{\circ}$ from the horizontal.  The image simulation is
effectively the ideal, infinite dose image of the perfect crystal
which provides a benchmark for both the low-dose experimental images
and our reconstructed image from the series.

\subsection{Variational Formulation of Pairwise Time Frame Registration}
\label{sec:Reg}
The usual single frame acquisition in STEM is replaced by a
statistical estimate of several low-dose short-exposure frames from
the same location of the specimen. By registering frames towards each
other and subsequent averaging we not only achieve better results by
diminishing spatial distortions but are also able to analyze beam
damage and its progress over time.

\subsubsection{Nomenclature and Mathematical Notation}
\label{subsec:MathNotation}
The scanning procedure creates an array of intensities that we call a
{\em frame}. It can be interpreted as an image {\em reflecting the
  properties} of the investigated specimen on some domain.  However,
due to distortions the locations corresponding to the measured
intensities are not known exactly.  Thus, the frame is not just the
image but is also related to an unknown positioning system determining
what exactly the image represents.  The question is how the
information received from several frames $f_j,\,j=1,\dots,n$, can be effectively combined
and how a specific knowledge about the distortions could be obtained.

To outline our approach we introduce some notation.  It is convenient
to represent a frame as a function $f$. The array values of the frame
define the function values $f(x)=f(x_1,x_2)$ for certain integer
coordinates $x_1$ and $x_2$ on a regular, rectangular grid.  Then,
using an appropriate approximation method, \eg bilinear
interpolation, one can extend this function to intermediate real
positions $x=(x_1,x_2)$.  Consider two frames $f_j$ and $f_k$.
We want to establish a correspondence between their positioning systems by
approximating the deformation $\phi_{j,k}$ that gives a position
$\phi_{j,k}(x)$ in frame $f_j$ for a position $x$ in frame $f_k$ in
the sense that $f_j\circ\phi_{j,k}\approx f_k$.  The process of
finding $\phi_{j,k}$ is often referred as \emph{registration} of $f_k$
and $f_j$.

\subsubsection{Rigid Registration}\label{subsect:rigid}
A particularly simple version of a registration map is a {\em rigid
  motion}, in general a combination of a translation and a rotation.
This is typically determined by first manually aligning the respective
frames, using specific image features, sometimes followed by
adjustments based on numerical indicators such as the
cross-correlation of the adjusted frames. In the case of STEM images,
collected pixel-by-pixel, rigid registration is inherently limited due
to the nature of the acquisition process.

\subsubsection{Overview of Solution for Distortion Map $\phi$}\label{subsect:Overview}

Our nonrigid series registration is founded on a variational approach that transforms two frames $f$ and $g$ into a common coordinate system with a nonparametric, nonrigid transformation $\phi$ such that $f\circ\phi\approx g$.
The objective functional consists of the sum of two terms, the \emph{normalized cross-correlation} of $f\circ\phi$ and $g$ as fidelity term, and the \emph{Dirichlet energy} of the displacement, \ie  $\phi$'s deviation from the identity as regularizer.
While this nonparametric transformation model is very flexible, an effective numerical realization is rather difficult due to the large number of local minima our objective functional exhibits. The proposed hybrid  minimization strategy rests on several important mathematical concepts:
\begin{itemize}
\item a multilevel scheme that processes the frames at different levels of resolution, going from coarse to fine;
\item minimization on each level based on a \emph{gradient flow},
 a generalization of the gradient descent concept;
\item an explicit time discretization of the gradient flow ODE combined with an automatic step size control to ensure convergence
\end{itemize}

\subsubsection{Nonrigid Registration}
\label{subsec:nonrigid_reg}
Since feature extraction from noisy data is a challenging task by
itself, we employ a registration approach that does not extract
features but works directly on the image intensities.  For a
comprehensive introduction to image registration in general we refer
the reader to the book by Modersitzki \cite{Mo04}.

Ideally, assuming no noise, no distortions, and a perfect extension of
the frames to real valued functions, one should expect that at any
$x=(x_1,x_2)$ from the scanned domain the composition $(f_j\circ
\phi_{j,k})(x)=f_j(\phi_{j,k}(x))$ has the same value as $f_k(x)$.  In
reality, one attempts to find a deformation $\phi_{j,k}$ that provides
a good fit of $f_j\circ \phi_{j,k}$ to $f_k$ by replacing the ideal
pointwise condition $(f_j\circ \phi_{j,k})(x)=f_k(x)$ with a suitable
\emph{similarity measure}. Such a measure quantifies the similarity of
two frames $f_j\circ \phi_{j,k}$ and $f_k$. An overview of various
popular similarity measures can also be found in Modersitzki~\cite{Mo04}.

To describe such a similarity measure, as usual, we denote by
$\overline f=\frac{1}{|\Omega|}\int_\Omega f \dx$ the mean value of $f$
over the image domain $\Omega$.  The standard deviation of $f$ over
$\Omega$ is denoted by $\sigma_f=\sqrt{\frac{1}{|\Omega|}\int_\Omega
  (f-\overline f)^2 \dx}$.  The normalized cross-correlation of two
functions $f$ and $g$ over the domain $\Omega$ is defined as
\[
\mathrm{NCC}[f,g]=\frac{1}{|\Omega|}\int_\Omega\frac{(f-\overline{f})}{\sigma_f}\frac{(g-\overline{g})}{\sigma_g} \dx.
\]
This quantity is between $-1$ and $1$, while $\mathrm{NCC}[f,g]=1$ if
and only if $f=a g+b$ for some real constants $a>0$ and $b$.  Adopting
the tradition to formulate a variational approach for determining the
deformation as a {\em minimization} problem, we define the data term
of our objective functional as
\begin{equation*}
 S[\phi]= S_{f,g}[\phi]:= -\mathrm{NCC}[f\circ\phi,g].
\end{equation*}
Without further constraints, finding a deformation
$\phi=(\phi^1,\phi^2)$ among all vector-valued piecewise bilinear
functions that minimizes $S[\phi]$ is a severely ill-posed problem.
Therefore, we add the regularization term
\begin{equation}
\begin{split}
R[\phi]&=\frac{1}{2}\int_\Omega \|D(\phi(x)-x)\|^2 \dx\\
 &= \frac{1}{2}\int_\Omega
 \left|\smallfrac{\partial\phi^1}{\partial x_1}-1\right|^2 \!\!\!+\!
 \left|\smallfrac{\partial\phi^1}{\partial x_2}  \right|^2 \!\!\!+\!
  \left|\smallfrac{\partial\phi^2}{\partial x_1} \right|^2 \!\!\!+\!
 \left|\smallfrac{\partial\phi^2}{\partial x_2}-1 \right|^2 \dx
\end{split}
\end{equation}
to penalize irregular displacements and receive the objective
functional
\begin{equation}
  \label{Ephi}
        E[\phi] = E_{f,g}[\phi] = S_{f,g}[\phi]+\lambda R[\phi]
\end{equation}
to be minimized. The larger the regularization parameter $\lambda>0$,
the smoother the resulting deformation $\phi$.  This regularization
term was used in Modersitzki\,\cite{Mo04} and
its use
is often referred to as {\em diffusion registration}.

While the non-rigid transformation model \eqref{Ephi} is very
flexible, its numerical realization is rather difficult.  Due to the
high beam sensitivity of zeolites, subsequent frames deteriorate
rapidly and the periodic nature of typical input frames renders the
estimation of the deformation $\phi$ that yields the global minimum of
$E[\phi]$ particularly challenging. In fact, even after including a
regularization term, the objective functional $E$ typically still
exhibits a large number of local minima.  Therefore, the numerical
minimization requires sophisticated algorithms designed to handle a
large number of local extrema. Hence, our proposed minimization
strategy rests on several conceptual pillars that guide the
development of the algorithmic procedures.  The remainder of the
current section provides the details of each component of the solution
process.  Primarily, it is based on a multilevel scheme that processes
different levels of resolution of the data to be registered (see {\bf
  Subsection~\ref{subsec:Multilevel_Numerical_Solver}}).  The
registration is first done on the coarsest level, the result obtained
on this level is prolongated to the next finer level and used as the
initialization for the minimization on this level.  The process is
repeated until a registration at the resolution of the input images is
achieved.  The minimization on each level is based on the gradient
flow concept which is described in {\bf
  Subsections~\ref{subsec:Gradient_Flow} and
  \ref{subsec:Numerical_discretization}}).

\subsubsection{Multilevel Numerical Solver for Variational Problem}
\label{subsec:Multilevel_Numerical_Solver}

A multilevel scheme serves as the outermost building block for the
registration of a pair of frames.  As indicated by numerous authors,
\eg \cite{ClDrRu02,Mo04,HaBeDrHoRuScSCUr07} just to name a few, a
coarse to fine registration strategy is a powerful tool to help
prevent a registration algorithm from getting stuck in undesired local
minima.  The basic ideas of multilevel schemes go back to multi-grid
algorithms \cite{Briggs,Hackbusch} and here we use the standard
terminology, such as prolongation and restriction mappings, from that
subject area.  In particular, a hierarchy of computational grids is
built, ranging from a very coarse grid level, \eg
$2^4\times2^4=16\times16$ pixels, to a fine level that matches the
resolution of the input images, \eg
$2^{10}\times2^{10}=1024\times1024$. The exponent of $2$ used to build
a grid in the hierarchy is called the grid level of the multiscale
grids. The registration is first done on a coarse staring level, the
result obtained on this level is prolongated to the next finer level
and used as the initialization for the minimization on this level. The
process is repeated until a registration at the resolution of the
input images is achieved. The coarser the level, the fewer structures
of the input images are preserved. Hence, the minimization at a
certain level avoids all local minima induced by the small structures
not visible at this level. Typically grids of the size $2^d\times2^d$
or $(2^d+1)\times(2^d+1)$ are used for multilevel schemes since there
are canonical prolongation and restriction mappings to transfer data
from one level to the other for these kind of grids.

In the case of $2^d\times2^d$ grids, the grid is interpreted in a
cell-centered manner. When going from one level to a finer level, each
grid cell is split into two times two cells (two in each coordinate
direction). The \emph{prolongation} of a function copies the value
from the coarse grid cell to the corresponding four grid cells in the
finer grid, while the restriction uses the average value of four grid
cells in the fine grid as the value for the corresponding coarse grid
cell. For $(2^d+1)\times(2^d+1)$ grids a mesh-centered approach is
used. Refining such a grid is done by adding a new grid node between
every pair of neighboring nodes in the coarse grid. The prolongation
copies the values of the coarse grid nodes to the nodes at the same
positions in the fine grid and determines the values for the fine grid
nodes that are not in the coarse grid by bilinear interpolation of the
neighboring coarse grid values. The \emph{restriction} is the
transposition of the prolongation operation, but rescaled row-wise
such that if we have a value of one on every fine grid node, we get a
value of one at every coarse grid node.

\subsubsection{Gradient Flow: Analytic Formulation}
\label{subsec:Gradient_Flow}

The minimization on a given level is based on a so-called
\emph{gradient flow}~\cite{SuYeMe07}, a generalization of the gradient
descent concept. As with the latter, the basic principle of the
minimization is to go in the direction of steepest descent and
formalized with the ordinary differential equation (ODE)
\begin{equation}
\label{eq:GradientFlowODE}
\partial_t\phi=-\grad_GE[\phi].
\end{equation}
Here, $\grad_G$ denotes the gradient with respect to a scalar product
$G$. In this notation, the well-known classical gradient in finite
dimensions is the gradient with respect to the Euclidean scalar
product. Since a gradient flow, as a gradient descent, is attracted by
the ``nearest'' local minimum, and the scalar product determines how
distances are measured, a suitable choice of the scalar product avoids
undesired local minima. As shown in \cite{ClDrRu02}, the ODE
\eqref{eq:GradientFlowODE} is equivalent to
\begin{equation}
\label{eq:GradientFlowODEReformulated}
\partial_t\phi=-A^{-1}E^\prime[\phi],
\end{equation}
where $E^\prime$ denotes the first variation of $E$ and $A$ is the
bijective representation of $G$, such that
$G(\zeta_1,\zeta_2)=\left<A\zeta_1,\zeta_2\right>$ for all
$\zeta_1,\zeta_2$. This formulation illustrates the influence of the
scalar product and leads us to choose it such that $A^{-1}$ is
smoothing its argument. This way the descent will avoid non-smooth
minimizers. A Sobolev $H^{1}$ inner product scaled by $\sigma$, \ie
\begin{equation}
\label{eq:SobolevInnerProduct}
G_\sigma(\zeta_1,\zeta_2)=\int_\Omega\zeta_1\cdot\zeta_2+\smallfrac{\sigma^2}{2}D\zeta_1:D\zeta_2\dx
\end{equation}
has proved to be a suitable choice. Here, we denote
$X:Y=\sum_{ij}X_{ij}Y_{ij}$ for matrices
$X,Y\in\mathbb{R}^{2\times2}$, \ie the Euclidean scalar product of the
matrices interpreted as long vectors. Note that for the corresponding
metric resulting from the inner
product~(\ref{eq:SobolevInnerProduct}), the operator $A^{-1}$ is
equivalent to one implicit time step of the heat equation with a step
size of $\smallfrac{\sigma^2}{2}$, an operation widely known for its
smoothing properties.

From \eqref{eq:GradientFlowODEReformulated} we see that the first
variation of the objective functional $E$ (\ie $E_{f,g}$) is necessary
in order to use the gradient flow. With a mostly straightforward but
somewhat lengthy calculation, one obtains
\begin{equation}
\label{eqn:weak_form}
\begin{split}
\var{E^\prime[\phi],\zeta} &= -\frac{1}{\abs{\Omega}}\int_\Omega
\brac{\nabla f(\phi)\cdot\zeta}\\
& \qquad\qquad\quad
\Brac{\smallfrac{\brac{g-\overline{g}}}{\sigma_{f\circ\phi}\sigma_g}+\smallfrac{\brac{f\circ\phi-\overline{f\circ\phi}}}{\brac{\sigma_{f\circ\phi}}^2}S[\phi]}\dx\\
& \quad
+\lambda\int_\Omega\brac{D\phi-\identity}:D\zeta\dx.
\end{split}
\end{equation}

\subsubsection{Gradient Flow: Numerical discretization}
\label{subsec:Numerical_discretization}
For the spatial discretization of a numerical solution to the weak
form (\ref{eqn:weak_form}) we employ piecewise bilinear Finite
Elements (FE) where the computational domain $\Omega$ is a uniform
rectangular mesh. Introducing the FE method is beyond the scope of
this paper, for a detailed introduction to the FE concept we refer to
the textbook by Braess \cite{Br07}. Of course, it is also possible to
use different discretization approaches, for instance Finite
Differences, but Finite Elements offer a canonical way to conveniently
handle $\var{E^\prime[\phi],\zeta}$, the weak formulation of the first
variation of $E$. To compute the integrals in the energy, its
variations and the metric, we use a numerical Gauss quadrature scheme
of order 3 (see, for example, \S 3.6 of Stoer and Bulirsch
\cite{StBu02}).

The gradient flow ODE \eqref{eq:GradientFlowODEReformulated} is solved
numerically with the Euler method, an explicit time discretization
approach. Thus, $\partial_t\phi$ is approximated by
$\smallfrac{\phi^{l+1}-\phi^l}{\tau}$, where $\phi^l$ denotes the
deformation at the $l$-th iteration of the gradient flow and $\tau$ is
the still-to-be-chosen step size. This leads to the update formula
\begin{equation}
\label{eq:DiscreteGradientFlow}
\phi^{l+1}=\phi^l-\tau A^{-1}E^\prime[\phi^l].
\end{equation}
Since the minimization does not require the knowledge of the
trajectory of $\phi$ given by the ODE but only the equilibrium point,
it is sufficient to use this kind of simple explicit method to solve
the ODE. It is very important though to carefully choose the step size
$\tau$ since the iteration can diverge with a poorly chosen step
size. To this end we employ the so-called \emph{Armijo rule with
  widening}~\cite{Ar66}. This rule is a line search method used on the
function
\[\Phi(\tau)=E\Brac{\phi^l-\tau A^{-1}E^\prime[\phi^l]}\]
and selects the step size such that the ratio of the secant slope
$\smallfrac{\Phi(\tau)-\Phi(0)}{\tau}$ and the tangent slope
$\Phi^\prime(\tau)$ always exceeds a fixed, positive threshold
$\rho\in(0,1)$. This means that the actual achieved energy decay
(secant slope) is at least $\rho$ times the expected energy decay
(tangent slope). Note that the specific choice of the step size
control is not important -- there are several popular step size
controls that could be used here. It is crucial though to use a step
size control that guarantees the convergence of the time discrete
gradient flow \eqref{eq:DiscreteGradientFlow}.

The resulting multilevel descent algorithm is summarized in {\sc Algorithm}~\ref{alg:MultilevelGradientFlow}.

\begin{algorithm}[h]
\caption{Multilevel gradient flow}
\label{alg:MultilevelGradientFlow}
\begin{algorithmic}
  \STATE{\textbf{Given} starting level $m_0$ and ending level $m_1$}
  \STATE{\textbf{Given} images $f=f^{m_1}$ and $g=g^{m_1}$}
  \STATE{\textbf{Given} initial deformation $\phi=\phi^{m_1}$}
  \FOR{$m=m_1-1,\ldots,m_0$}
    \STATE{Initialize $[f^{m}, g^{m}, \phi^{m}]$ by restricting $[f^{m+1}, g^{m+1}, \phi^{m+1}]$}
  \ENDFOR

  \FOR{$m=m_0,\ldots,m_1$}
    \STATE{Register $f^m$ and $g^m$ using the gradient flow (cf.~\eqref{eq:DiscreteGradientFlow})\par on level $m$, with $\phi^{m}$ as initial deformation}
    \STATE{Store the resulting deformation in $\phi^{m}$}
    \IF{$m<m_1$}
      \STATE{Set $\phi^{m+1}$ to the prolongation of $\phi^{m}$}
    \ENDIF
  \ENDFOR
  \STATE{\textbf{return} $\phi^{m_1}$}
\end{algorithmic}
\end{algorithm}

Our multilevel descent algorithm is well-suited for the beam sensitive
materials we consider here.  Beam damage starts as
a loss of high frequency information in the image as the atomic
structure of the sample is lost with increasing electron irradiation.
At the coarser stages of the registration, we find the distortion map
for the large scale differences between the image frames.  As the
registration proceeds, the distortions for the earlier distortion map
are used as the initial values for the next level.  In the case where
either the noise level in the input frames is extremely high or the
material has changed from frame to frame, the distortions from the
coarser level are retained and the regularization of the algorithm
smoothly adjusts between the coarser grid points. The step size
control described above means that our registration proceeds only to
the point where the comparison between the frames is still
meaningful.  When either the noise level of the input or a significant
difference in the two frames occurs, the registration algorithm is
terminated, minimizing the artifacts of our method.

\subsection{Registration of a frame series}
\label{sec:si:FrameSeriesRegis}

Now that we have proposed how to register a pair of images, we still
have to provide a concept on how to register a series of images when
the goal is to fuse the information contained in all images of the
series into a single image. The basic idea here is to register all of
the input frames to a common reference frame. Once this is achieved
one can simply calculate a suitable average of the registered images
pixel by pixel to get the reconstructed image.

A natural strategy here is to simply select one of the images as
reference frame and then to use the registration method for image
pairs to register all of the other images to the selected
image. Figure~\ref{fig:si:ZeoliteInputFrames} indicates some of the
potential pitfalls of this strategy by showing two frames of a series
that are several frames apart. The left image depicts the first frame
of the series, the right image depicts the ninth frame. Most
strikingly, the specimen moved considerably between the two
frames. Moreover, the thin edge region of the specimen appears
differently in both images. The material is so beam sensitive that the
scanning already caused visible damage to part of the specimen in the
latter image. In combination with the periodic structure of the
material these effects increase the difficulty when registering this
image pair.  A good initial guess for the registration where the
spatial error is already less than half of the unit cell size of the
specimen should be sufficient to overcome these problems.  In order to
obtain such an accurate initialization we can exploit properties of
the acquisition process of the series (see {\bf
  Subsection~\ref{subsec:Basic_Strategy_Series}} for details).

The main procedure for registering a series is summarized by the
steps
\begin{itemize}
\item use of the identity as an initial guess only when registering \emph{consecutive} frames and subsequently employ the composition of the already established transformations for this purpose, since the specimen can move considerably during the series' acquisition;
\item register the series to a single reference frame (usually the first)
and calculate a suitable pixel-wise average of the registered series to fuse the information contained in all frames into a single one;
\item iterate this procedure with lowered weight on the regularizer using the average as new common reference frame.
\end{itemize}
which are fully described in the following subsections.

\subsubsection{Basic Strategy for a Series}
\label{subsec:Basic_Strategy_Series}
Since consecutive images in our input series were acquired directly
one after another, we can assume the differences between two
consecutive images, both regarding displacement and beam damage, to be
small enough to use the identity as initialization for the
minimization on the coarsest level in
Algorithm~\ref{alg:MultilevelGradientFlow}. In case this assumption is
not fulfilled it may be necessary to add a fiducial mark to the
specimen to facilitate the registration algorithm.

\begin{figure}[ht]
\begin{center}
  \resizebox{0.7\linewidth}{!} {
 \begingroup
  \makeatletter
  \providecommand\color[2][]{%
    \errmessage{(Inkscape) Color is used for the text in Inkscape, but the package 'color.sty' is not loaded}
    \renewcommand\color[2][]{}%
  }
  \providecommand\transparent[1]{%
    \errmessage{(Inkscape) Transparency is used (non-zero) for the text in Inkscape, but the package 'transparent.sty' is not loaded}
    \renewcommand\transparent[1]{}%
  }
  \providecommand\rotatebox[2]{#2}
  \ifx\svgwidth\undefined
    \setlength{\unitlength}{202.03199463pt}
  \else
    \setlength{\unitlength}{\svgwidth}
  \fi
  \global\let\svgwidth\undefined
  \makeatother
  \begin{picture}(1,0.29278248)%
  \ifpdf
    \put(0,0){\includegraphics[width=\unitlength]{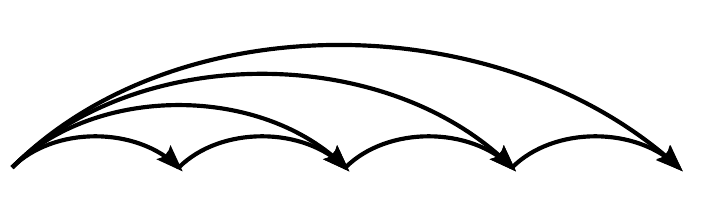}}%
    \put(-0.00128686,0.0071747){\color[rgb]{0,0,0}\makebox(0,0)[lb]{\smash{$f_1$}}}%
    \put(0.23629928,0.0071747){\color[rgb]{0,0,0}\makebox(0,0)[lb]{\smash{$f_2$}}}%
    \put(0.47388542,0.0071747){\color[rgb]{0,0,0}\makebox(0,0)[lb]{\smash{$f_3$}}}%
    \put(0.71147152,0.0071747){\color[rgb]{0,0,0}\makebox(0,0)[lb]{\smash{$f_4$}}}%
    \put(0.94905765,0.0071747){\color[rgb]{0,0,0}\makebox(0,0)[lb]{\smash{$f_5$}}}%
    \put(0.49368426,0.26455967){\color[rgb]{0,0,0}\makebox(0,0)[lb]{\smash{$\phi_{5,1}$}}}%
    \put(0.09770737,0.04677239){\color[rgb]{0,0,0}\makebox(0,0)[lb]{\smash{$\phi_{2,1}$}}}%
    \put(0.33529349,0.04677239){\color[rgb]{0,0,0}\makebox(0,0)[lb]{\smash{$\phi_{3,2}$}}}%
    \put(0.57287964,0.04677239){\color[rgb]{0,0,0}\makebox(0,0)[lb]{\smash{$\phi_{4,3}$}}}%
    \put(0.81046574,0.04677175){\color[rgb]{0,0,0}\makebox(0,0)[lb]{\smash{$\phi_{5,4}$}}}%
    \put(0.59267848,0.16556545){\color[rgb]{0,0,0}\makebox(0,0)[lb]{\smash{$\phi_{4,1}$}}}%
    \put(0.35509234,0.14576661){\color[rgb]{0,0,0}\makebox(0,0)[lb]{\smash{$\phi_{3,1}$}}}%
  \else
    \put(0,0){\includegraphics[width=\unitlength]{Figure_3}}%
  \fi
  \end{picture}%
\endgroup
   }
\end{center}
\caption{Strategy to match a series of consecutive images.}
\label{fig:si:seriesMatchingStrategy}
\end{figure}

After having a suitable deformation for each consecutive image pair we
can chain those to get estimates for non-consecutive pairs. Let
$f_1,\ldots f_n$ denote the images in our series ordered by their time
of acquisition.  As outlined above we can estimate deformations
$\phi_{2,1}$ and $\phi_{3,2}$ that match the first and second image
pair respectively, \ie $f_2\circ\phi_{2,1}\approx f_1$ and
$f_3\circ\phi_{3,2}\approx f_2$. Then $\phi_{3,2}\circ\phi_{2,1}$ is a
good initial guess for the deformation matching $f_1$ and $f_3$
because
$f_3\circ(\phi_{3,2}\circ\phi_{2,1})=(f_3\circ\phi_{3,2})\circ\phi_{2,1}\approx
f_2\circ\phi_{2,1}\approx f_1$. Therefore, our proposed registration
algorithm with $\phi_{3,2}\circ\phi_{2,1}$ as initial guess allows us
to reliably register $f_3$ to $f_1$. Iterating this provides a way to
accurately register any frame of the series to the first frame
(cf. Figure~\ref{fig:si:seriesMatchingStrategy}).

After registering all frames to the first one, we can fuse the information contained in the series by a suitable averaging operator ${\cal A}$ acting on the deformed frames, $g_i=f_i\circ\phi_{i,1}$,
where $\phi_{1,1}$ is the identity mapping.
Given a sequence of frames $\{g_i\}_{i=1}^{n}$ the first  example is
the mean
\begin{equation}
\label{eq:si:arithmeticMean}
{\cal A}[\{g_i\}_i](x)=\frac{1}{n}\sum_{i=1}^n g_i(x)
\end{equation}
which provides the best least squares fit to the sequence.  The
arithmetic mean is a natural choice to calculate an average and is
very easy to compute, although, depending on the input data, it may
not be the best choice.  In particular, if the input data contains
noticeable outliers, which due to the scattering nature of the
electrons in the HAADF STEM measurements is expected to be the case,
the {\em median} is preferable to the arithmetic mean.  Recall the
definition of the median
\begin{equation}
\label{eq:si:median}
{\cal A}[\{g_i\}_i](x)\in\argmin_{g\in\mathbb{R}}\sum\limits_{i=1}^n\abs{g_i(x)-g},
\end{equation}
which is the best $\ell_1$ fit to the sequence and therefore less sensitive to outliers.
This selection of ${\cal A}$ is used in all subsequent experiments.

\begin{figure*}[htp]
\begin{center}
\begin{tabular}{cc}
{\begin{overpic}[width=0.475\linewidth]{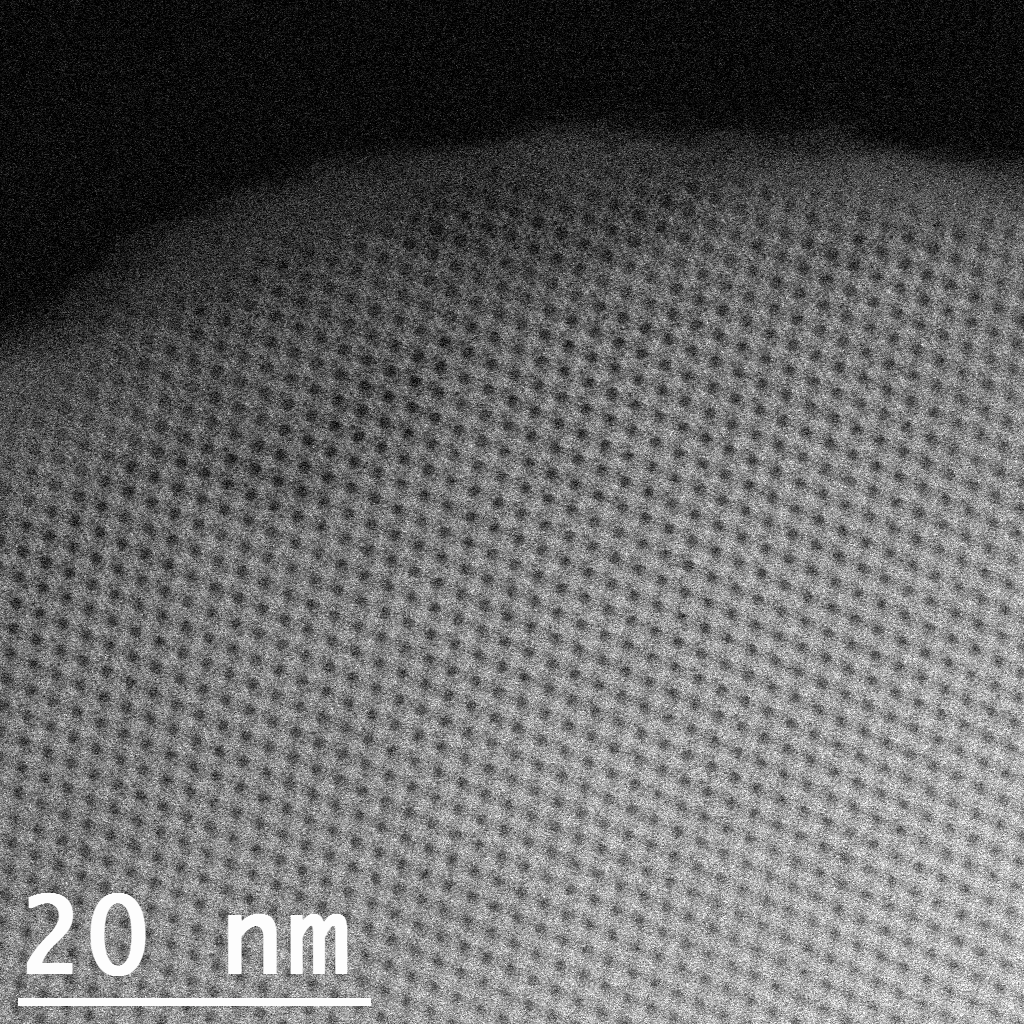}
  \put(2,94){\colorbox{white}{\Large a}}\end{overpic}}&
{\begin{overpic}[width=0.475\linewidth]{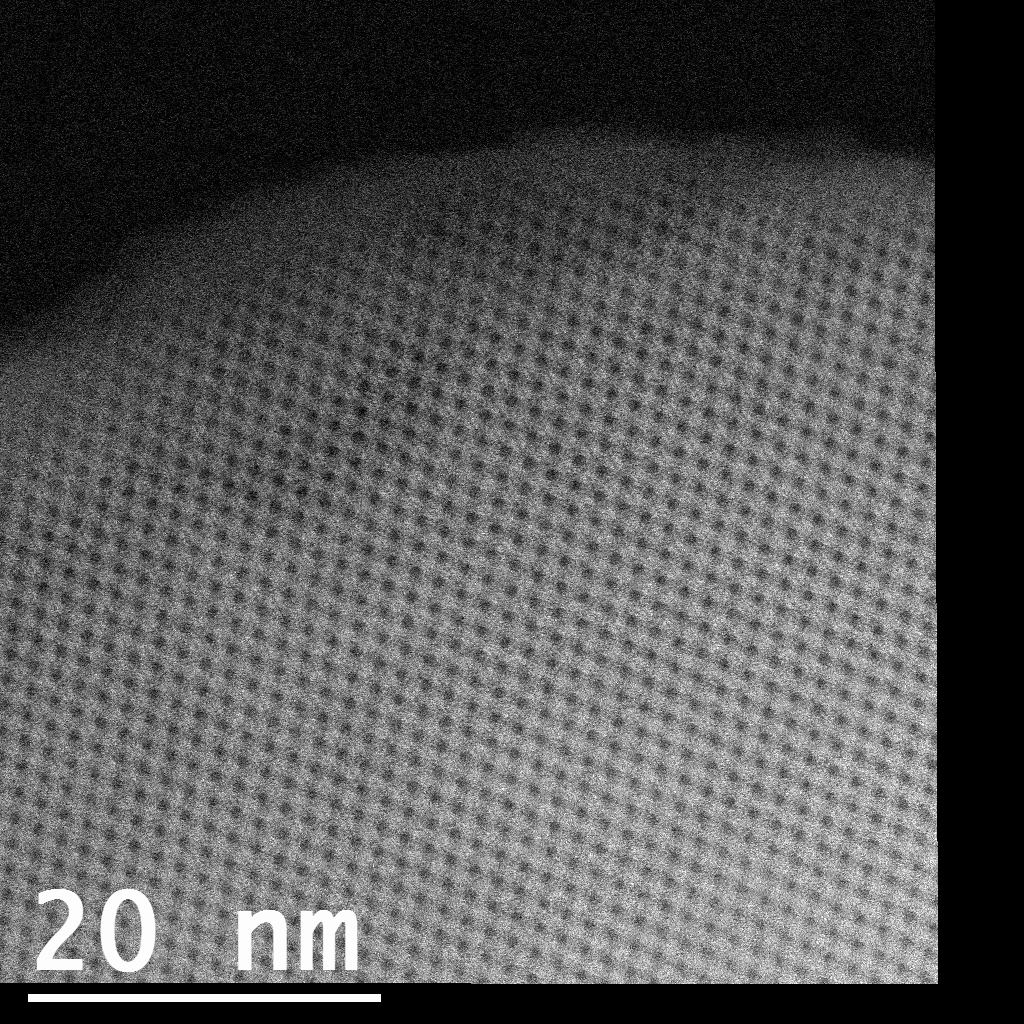}\put(2,94){\colorbox{white}{\Large b}}\end{overpic}}
\end{tabular}
\caption{
{\bf Registration of the raw data.} The raw data of the first frame of
a HAADF STEM image series {\bf (a)}  and the ninth frame {\bf(b)} following the non-rigid registration procedure.
The solid black at the bottom and right edge of the registered image are parts of the field of view from the first frame, and used in the reconstruction from the full image series, which do not appear in the ninth frame.
}
\label{fig:ninthDeformed}
\end{center}
\end{figure*}

\subsubsection{Iteration of Basic Strategy}
\label{subsec:iteration_series}
The estimated reconstructed image $f={\cal A}\{f_i\circ\phi_{i,1}\}$
from the previous section can be considerably improved by iteration of
this basic procedure.  Choosing $f_1$ as reference frame was a
canonical but arbitrary choice, whereas the newly obtained average $f$
is a much more suitable reference frame. Thus, it is reasonable to
repeat the process using $f$ as reference frame. Since $f_1$ was the
reference frame when calculating $f$ the identity mapping is a good
initialization for the registration of $f_1$ to $f$ and we can simply
prepend $f$ to the input series, calling it $f_0$, and repeat the
process. Here however, we don't include $f_0$ when calculating the
average since the original images $f_1,\ldots f_n$ obviously contain
all of our measurement information. Thus, an average of the deformed
frames may be calculated via $f={\cal A}[\{f_i\circ\phi_{i,0}\}_i]$.
This results in a further improved reconstructed image $f$ and thus
the process can be repeated again, leading to an iteration procedure
to calculate $f$.

The $k$-th estimate of $f$ is denoted by $f^k$. Initially,
$f^0=f_1$ since the specimen is least damaged during the acquisition
of $f_1$. We use $f_0=f^{k-1}$ as our reference frame at this stage
and calculate $f^{k}$ using ${\cal A}[\{f_i\circ\phi_{i,0}^k\}_i]$ where $\phi_{i,0}^k$ denotes the deformation from $f_i$ to $f_0=f^{k-1}$
at the end of each iteration.

A critical issue in finding the deformations $\phi_{\changed{i},0}^k$ is how to
determine the initial guess for the nonrigid registration
process we have described.
Since the specimen can move considerably during the
acquisition of the image series, we use the identity as an initial
guess only for the transforms $\phi_{j+1,j}$ between two consecutive
frames (note that $\phi_{1,0}^0$ is the identity).  Then the initial
guess for $\phi_{1,0}^{k+1}$ is the transform $\phi_{1,0}^{k}$, while
for $\phi_{j+1,0}^k$ it is the composition
$\phi_{j+1,j}^k\circ\phi_{j,0}^k$. Note that the estimation of
$\phi_{j+1,0}^k$ depends on
$\phi_{j,0}^k$, so as indicated before, these deformations need to be determined one
after another. As an example, in Figure~\ref{fig:ninthDeformed} we present
frame~9 registered to frame~1.

This core methodology opens an avenue to further algorithmic options which facilitate an even increased
denoising effectiveness at the expense of additional computational effort (see {\bf Subsection~\ref{subsec:extended_series}}).
The point is that within some regime, in principle,  additional computational work gains increased information quality.
The full strategy is summarized in {\sc Algorithm}~\ref{alg:SeriesAveraging}.
\begin{algorithm}
\caption{Series averaging procedure}
\label{alg:SeriesAveraging}
\begin{algorithmic}
  \STATE{\% \textit{Register each consecutive input image pair}}
  \FOR{$i=2,\ldots,n$}
    \STATE{Compute $\phi_{i,i-1}$ with Algo.~\ref{alg:MultilevelGradientFlow} (initial guess identity)}
  \ENDFOR
  \STATE{}
  \STATE{Initialize $0$-th average with $f^0:=f_1$}
  \FOR{$k=1,\ldots,K$}
    \STATE{Select last average as reference frame, \ie $f_0:=f^{k-1}$}
    \STATE{\% \textit{Registration part}}
    \STATE{Compute $\phi^k_{1,0}$ with Algo.~\ref{alg:MultilevelGradientFlow} (initial guess identity)}
    \FOR{$i=2,\ldots,n$}
      \STATE{Compute $\phi^k_{i,0}$ with Algo.~\ref{alg:MultilevelGradientFlow} (initial guess $\phi_{i,i-1}\circ\phi^k_{i-1,0}$)}
    \ENDFOR
    \STATE{}
    \STATE{\% \textit{Averaging part}}
    \STATE{Obtain average $f^k$ pixel-wise via \eqref{eq:si:arithmeticMean} or \eqref{eq:si:median}}
  \ENDFOR
\end{algorithmic}
\end{algorithm}

Note that after the first time the average has been computed, \ie when
$k\geq2$, the reference frame $f_0$ is considerably less noisy than
the input data. Thus, it is then possible to reduce the regularization
parameter $\lambda$ when calculating $\phi^k_{i,0}$ for
$k\geq2$. Here, we typically use $0.1\,\lambda$ as regularization
parameter for $k\geq2$, \ie the same regularization parameter value is
used for all $k\geq2$, but this value is smaller than the one used for
$k=1$.

\section{Result and Discussion}
\begin{figure*}[htbp]
  \begin{center}
    \begin{tabular}{cc}
      \begin{overpic}[width=0.475\linewidth]{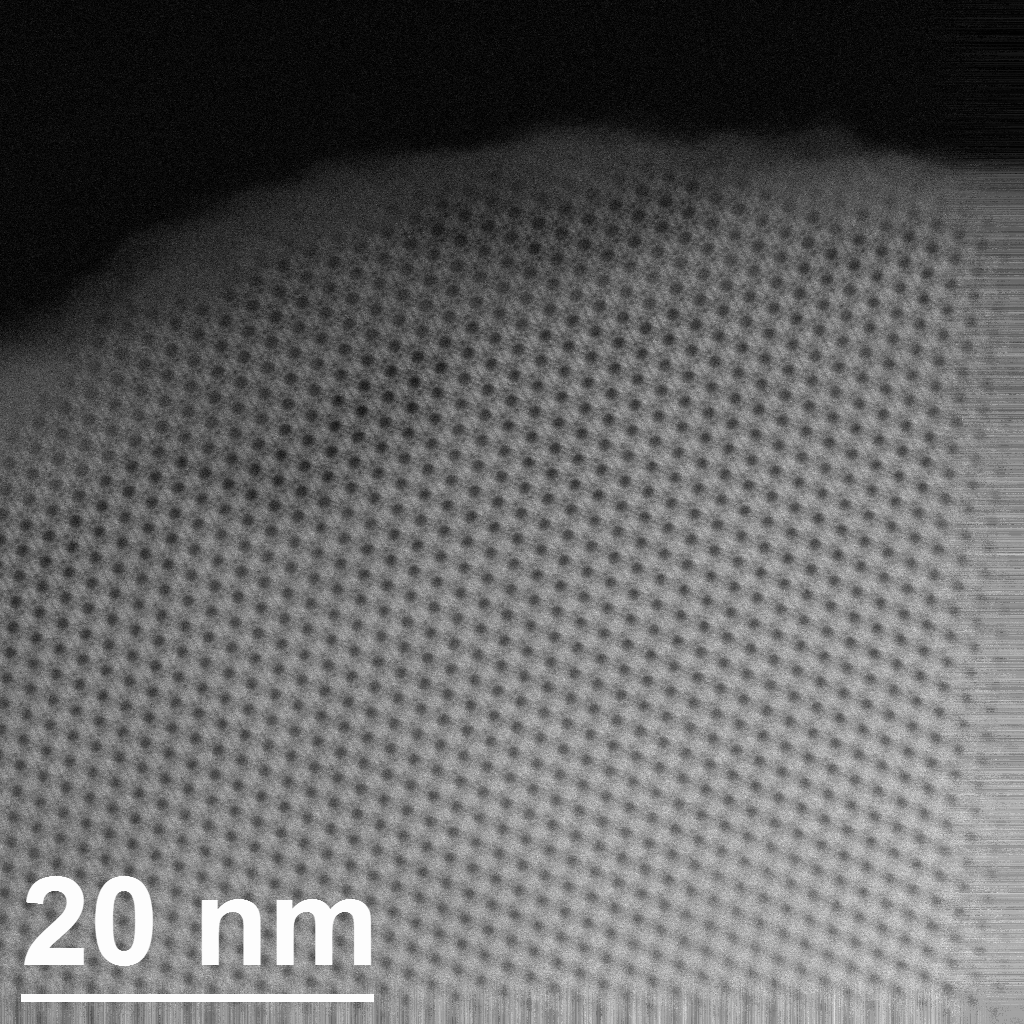}
        \put(2,94){\colorbox{white}{\Large a}}\end{overpic}&
      \begin{overpic}[width=0.475\linewidth]{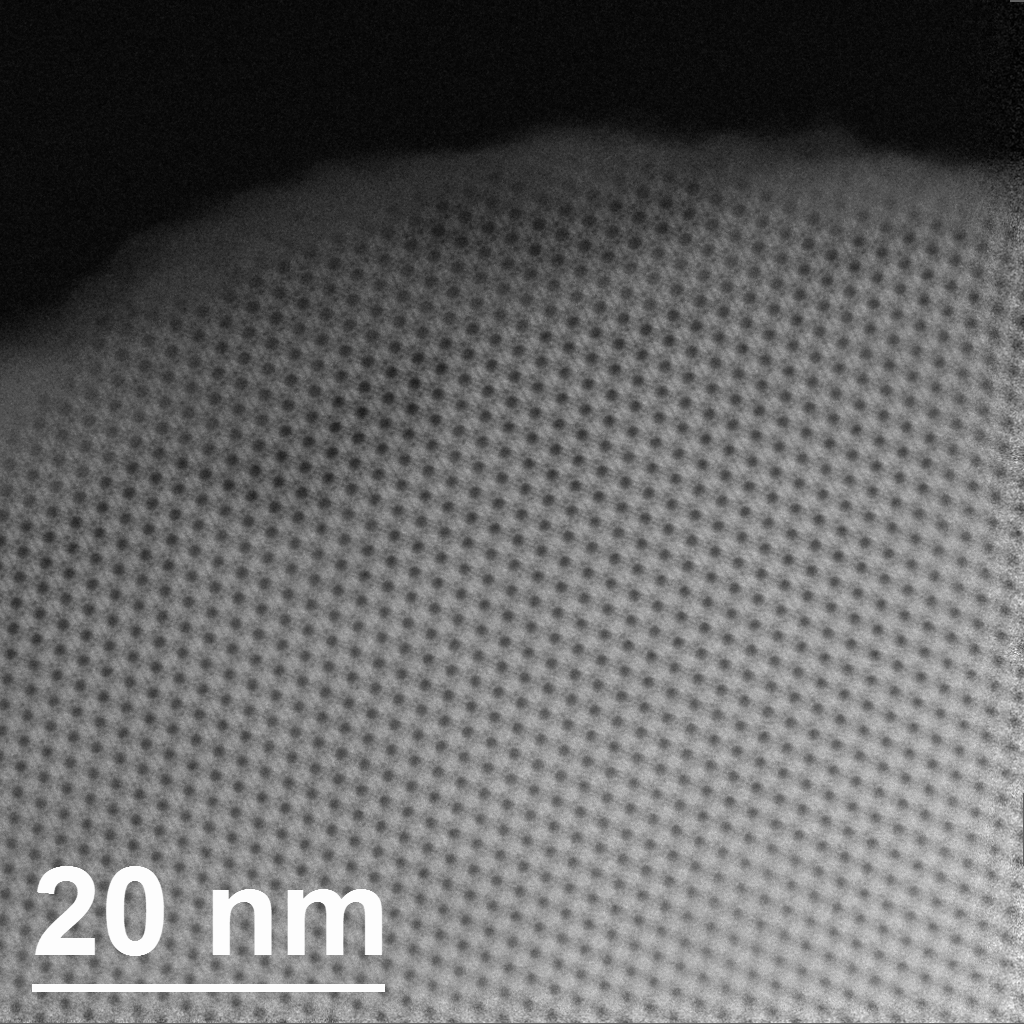}
        \put(2,94){\colorbox{white}{\Large b}}\end{overpic}
    \end{tabular}
    \caption{ {\bf High resolution, full frame results of manual and
        nonrigid registration.}  Resulting average images produced by
      registration of 9 frames for a) manual  and b)nonrigid
      registration, followed by median averaging.  (Images are
      embedded in the on-line PDF at full resolution.)
    }
    \label{fig:high-res_compare}
  \end{center}
\end{figure*}
An application of the algorithm to our test series of nine
frames produces the image~b) in the Figure~\ref{fig:high-res_compare}
below.  For comparison, a manual, rigid alignment of the same input
frames was performed and median averaging applied.  The resulting
image is shown in Figure~\ref{fig:high-res_compare}a. The non-rigid
algorithm required slightly over 2 hours of computation on an Intel
i5-3335s@2.7GHz for an average time to register two 1024$\times$1024
frames of 3.66 minutes.

\begin{figure}[htb]
  \begin{center}
    \begin{tabular}{cc}
      \begin{overpic}[width=0.47\linewidth]{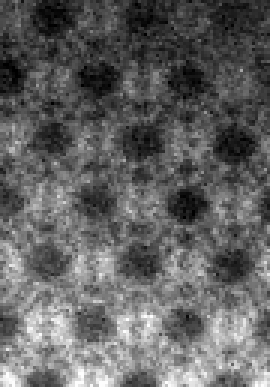}
        \put(2,94){\colorbox{white}{\Large a}}\end{overpic}&
      \begin{overpic}[width=0.47\linewidth]{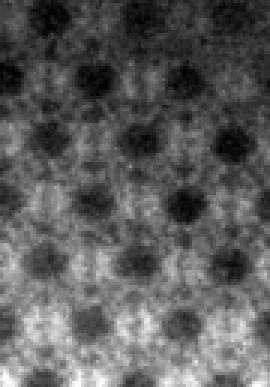}
        \put(2,94){\colorbox{white}{\Large b}}\end{overpic}
  \end{tabular}
\end{center}
\caption{{ {\bf Results of frame assimilation.} Segments of the
    assembled images produced by the two methods of registration of 9
    frames, followed by median averaging. The displayed region is
    equivalent to that of the input frame seen in
    Figure~\ref{fig:si:multislice}.  ~~{\bf a)} Resulting image
    segment from {\em manual rigid registration} of 9~image frames.
    ~{\bf b)} An image segment produced with the aid of a {\em
      non-rigid registration} of the 9~image frames.  Super-cages and
    the double six-member rings are resolved across this entire
    segment, as they are in the full image seen in
    Figure~\ref{fig:high-res_compare}.  } }
\label{fig:six}
\end{figure}

\subsection{Metric for Success of Reconstruction.}
\label{sec:ReconstructionSuccessMetric}
Figure~\ref{fig:six} shows a magnified portion of
Figure~\ref{fig:high-res_compare} for both the manual rigid
registration as well as the algorithm described above. The reference
area for the manual alignment, near the material thin edge, is
included in this magnified segment, which represents the best-case
results of the rigid alignment. The image in Figure~\ref{fig:six}\,b
exemplifies the successful implementation of the described paradigm
for a series of nine HAADF STEM images of the Si-Y zeolite sample. In
particular, this illustrates the effect of the algorithm for beam
sensitive materials.  The improvement offered by the computational
procedure is visually apparent as even the secondary pores become
visible essentially over the complete image area covered by the
material.  Figure~\ref{fig:si:ZeoliteInputFrames} shows the
substantial translation undergone by the specimen during the
acquisition process. How to factor out such a translation from the
series of frames is outlined in the {\bf
  Section~\ref{sec:si:BasicDistortionModel}} which could further
improve the registration quality.

\subsection{Quantitative Measures of Quality}
As important as a complete robust and accurate information processing
chain is a {\em quantitative} quality assessment that can serve as a
basis for subsequent scientific conclusions. The following discussion
should be viewed as a step in this direction. One aspect is to concede
that perhaps a ``single number'' as ``measure of quality'' can hardly
serve our purpose. In view of the complexity of the information
carried by the data, as the comparison of rigid versus nonrigid
registration shows, one should rather incorporate {\em localizing}
measures, examples of which are suggested below.

In order to quantify the quality of the registration, we consider one
type of
{\em information content} remaining in the
quantity
\begin{equation}
\label{eq:Difference}
d=f_n-f_0\circ \phi_{0,n},
\end{equation}
\ie the difference between
the last experimentally acquired frame $f_n$ and our average $f_0$
nonrigidly transformed back to $f_n$. Since the algorithm we
have described in the previous subsection
calculates $\phi_{n,0}$ instead of $\phi_{0,n}$, at
first glance it might seem natural to consider the difference
$f_0-f_n\circ\phi_{n,0}$.  However, this comparison would be flawed
because the experimentally acquired data $f_n$ would then have been
manipulated before being compared
to the calculated reconstruction.  To calculate $\phi_{0,n}$ we simply
register our average $f_0$ to $f_n$ with Algorithm~\ref{alg:MultilevelGradientFlow},
where a numerical inverse of the already calculated $\phi_{n,0}$ is
used as an initial guess.
Note that to prevent any smoothing in the pixel-wise calculation of $f_0\circ\phi_{0,n}$, $f_0$ is evaluated at the deformed pixel positions using nearest neighbor interpolation, not bilinearly like in the FE-based registration algorithm.

In case $f_0$ is the noiseless, undistorted image we aim to reconstruct and $\phi_{0,n}$ perfectly captures the distortion of the target image $f_0$ to the $n$-th measured frame $f_n$, $d$ contains nothing but noise.
Thus, the higher the quality of the registration and averaging procedure,
the closer $d$ is to pure noise.  Moreover, in order to compare different
methods one can consider a measurement that evaluates the amount of
information in $d$.  Of course, in our particular example where the
frames change due to accumulation of beam damage during the
acquisition $d$ will contain the induced changes between the frames.
Since the induced change is primarily the loss of high frequency
information, it will be difficult to distinguish between the high
frequency noise in the frames due to the low-dose nature of the
acquisition.

\begin{figure*}[ht]
  \begin{center}
    \begin{tabular}{ccc}
      \begin{overpic}[width=0.31\linewidth]{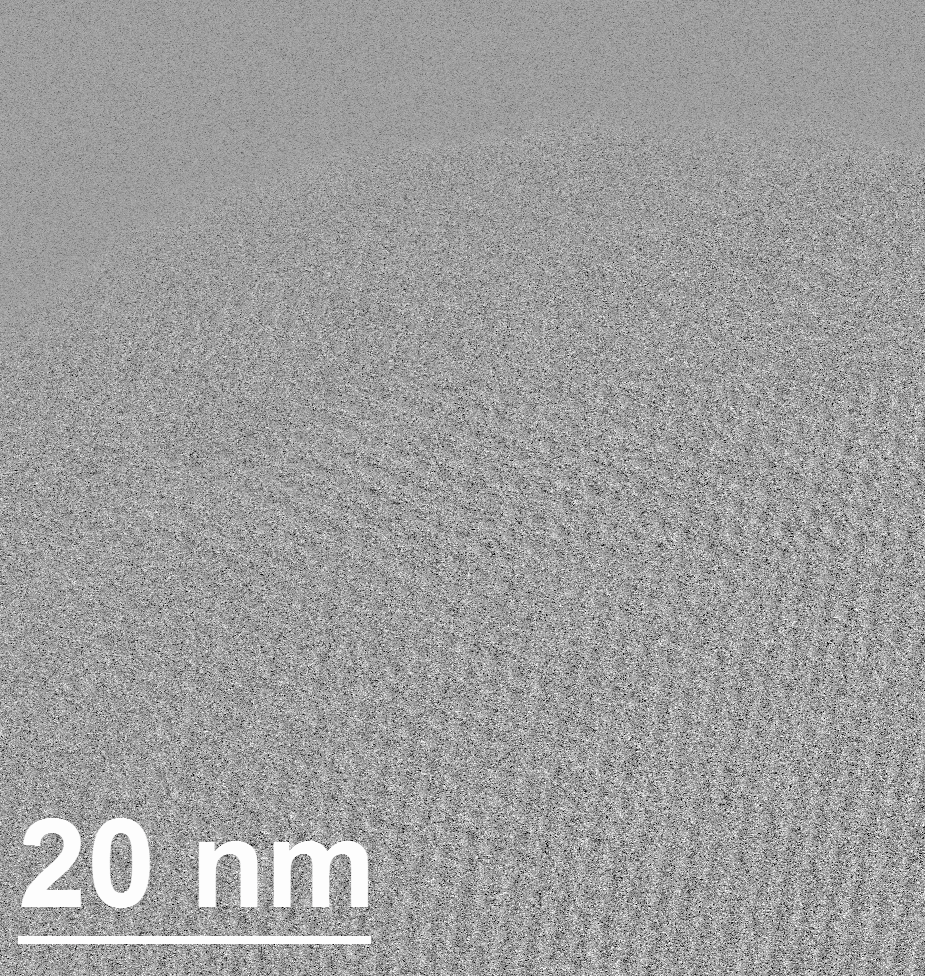}
        \put(2,94){\colorbox{white}{\Large a}}\end{overpic}&
      \begin{overpic}[width=0.31\linewidth]{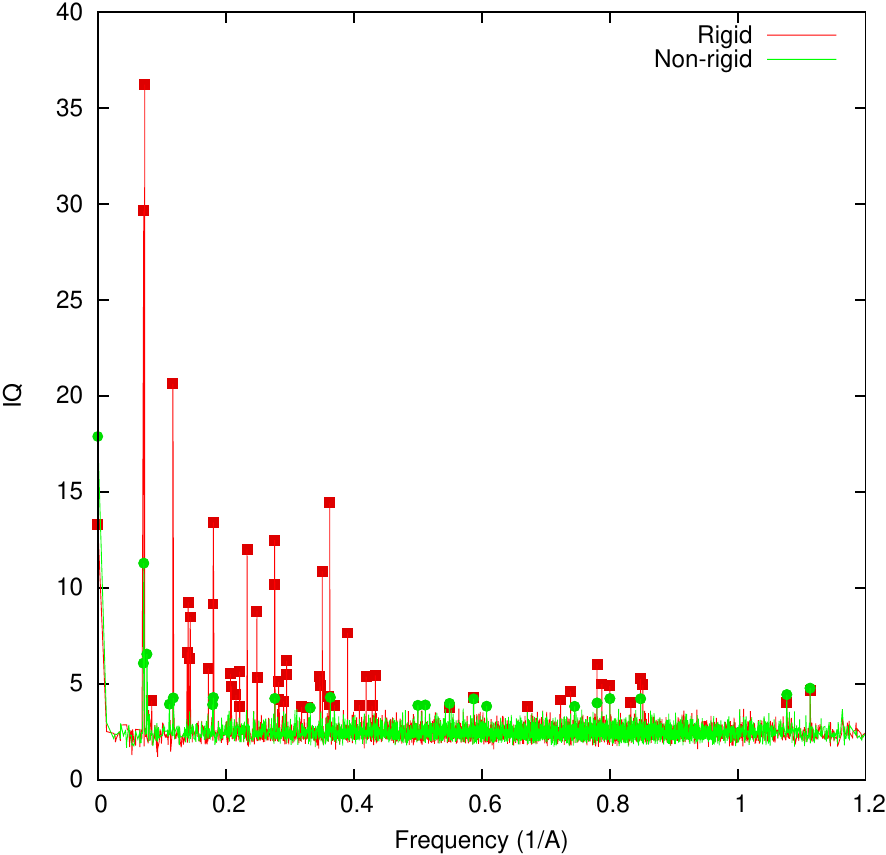}
        \put(2,100){\colorbox{white}{\Large c}}\end{overpic}&
      \begin{overpic}[width=0.31\linewidth]{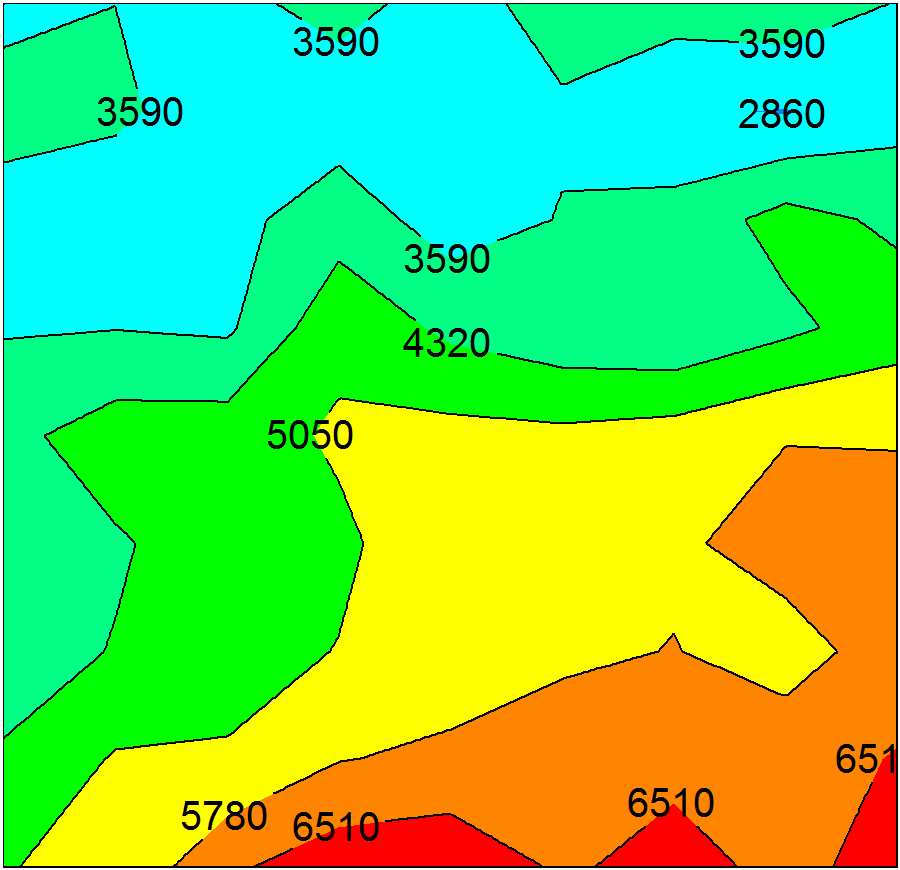}
        \put(2,101){\colorbox{white}{\Large e}}\end{overpic}\\[1.5ex]
      \begin{overpic}[width=0.31\linewidth]{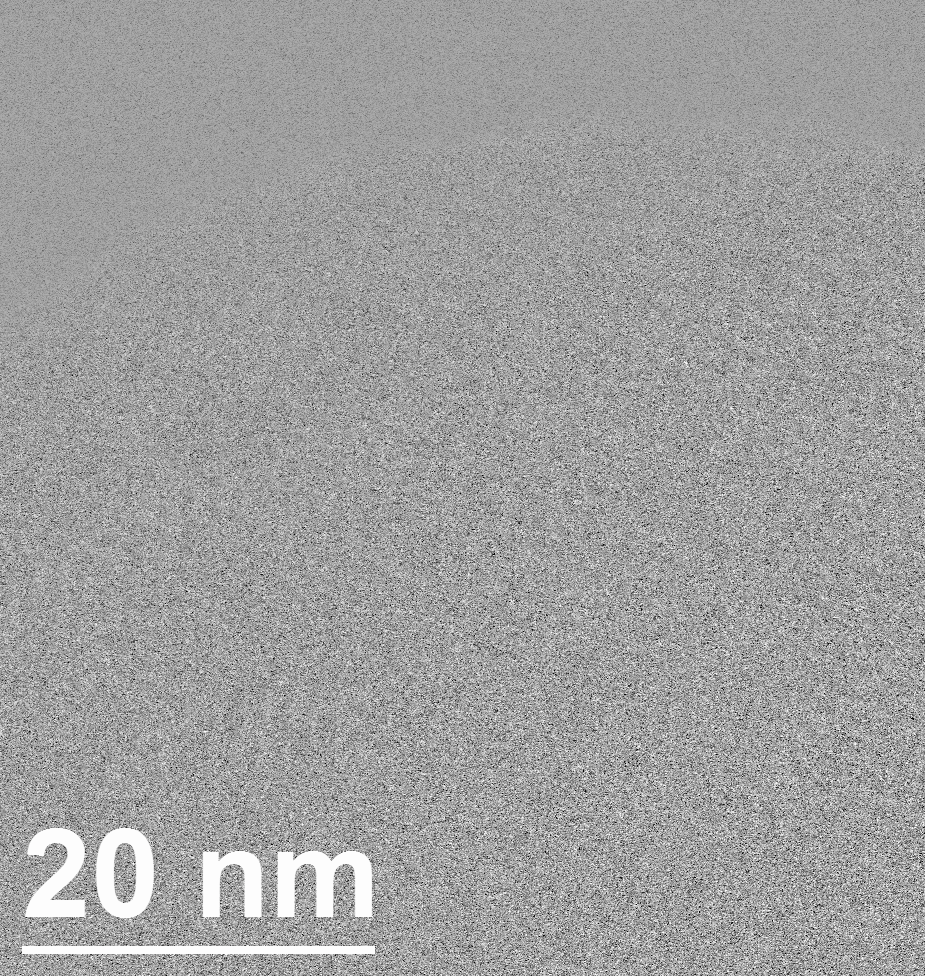}
        \put(2,93){\colorbox{white}{\Large b}}\end{overpic}&
      \begin{overpic}[width=0.31\linewidth]{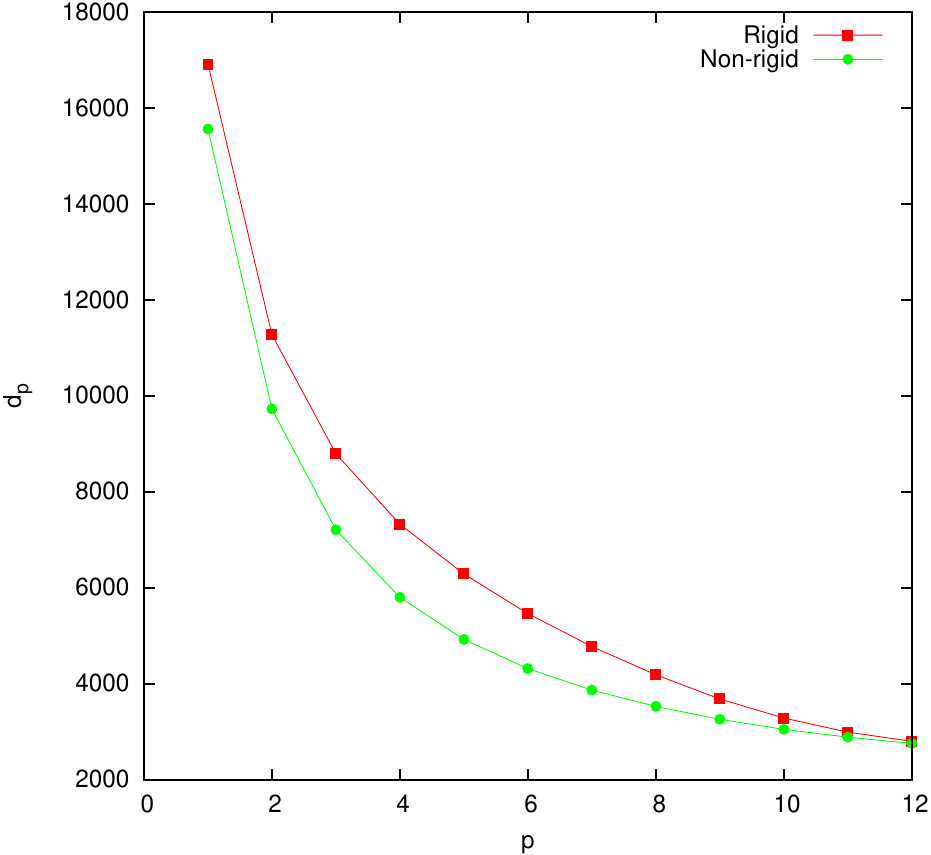}
        \put(2,98){\colorbox{white}{\Large d}}\end{overpic}&
      \begin{overpic}[width=0.31\linewidth]{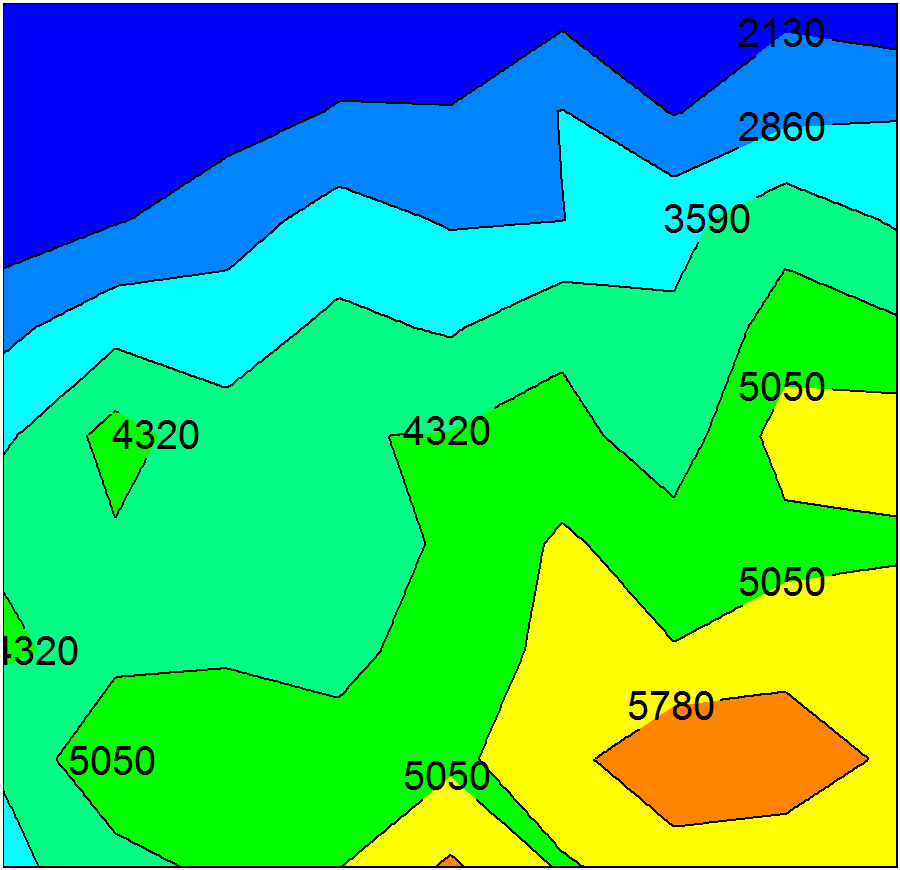}
        \put(2,99){\colorbox{white}{\Large f}}\end{overpic}
    \end{tabular}
 \caption{{\bf Quality of Reconstructions.}
 ~The left column of images in the figure show the difference
$d=f_0\circ\phi_{0,9}-f_9$ evaluated at the deformed pixel positions using nearest neighbor
interpolation,~{\bf a)} for the manual rigid registration and {\bf b)} for the automatic non-rigid registration, cropping as needed.
In the case of perfect alignment one would expect only noise in these images.
{\bf c)}~IQ plots of images~a) (red) and~b) (green) plotted against distance from the origin. Smaller IQ values indicate smaller residual
information in an image.
~{\bf d)}~Arithmetic mean of $d_p$ (as the patch radius $p=1,\ldots12$
varies)  for the manual rigid (red) and the automatic non-rigid
registration (green) of the first nine frames of a HAADF STEM time
series, where the median was used to calculate the average image.~{\bf
  e)}~Arithmetic mean of $d_2$ for $9\times 9$ subregions for the
manual rigid registration.~{\bf f)}~Arithmetric mean of $d_2$ for
$9\times 9$ subregions for the automatic non-rigid registration.
\label{fig:quality_compare}
 }
\end{center}
\end{figure*}

We define such measurements in the next subsection.
Figure~\ref{fig:quality_compare} summarizes
the performance of the manual and the nonrigid
registration methods based on these
procedures. Figure~\ref{fig:quality_compare}a is $d$ for the manual, rigid
registration, while Figure~\ref{fig:quality_compare}b is $d$ for the
non-rigid registration.  For the manual rigid alignment, a small
portion of the image consists of noise, while the majority of the
frame contains significant structural information which will be
blurred upon averaging the frames.  The non-rigid registration
approaches the ideal of pure noise throughout the entire frame.

\subsection{IQ Factor for estimating Image Quality.}
In the case that $d$ is identical to noise, the power spectrum of $d$
should contain no maxima in its modulus.  Evaluating the power
spectrum of $d$ will then provide a measurement of the quality of the
registration.  The IQ-factor introduced by Unwin and
Henderson~\cite{UnwinHenderson} provides a well known method of
measuring the local maxima in the modulus of a power spectrum of an
image.  The IQ-factor is the ratio of the maximum of the modulus of a
spot in the power spectrum to the average local background in the
power spectrum. In this case, since we are interested in measuring the
error in the registration process by quantifying how close $d$ is to
pure noise, a low IQ indicates excellent registration while a high
IQ-factor indicates failure in the registration process.  Note that
this is inverse to the usual meaning of the IQ-factor.
Figure~\ref{fig:quality_compare}\,c shows how $d$ for the manual rigid
alignment of the HAADF STEM image series
(Figure~\ref{fig:quality_compare}\,a) contains a significant amount of
non-noise information with a maximum IQ of more than 35.  In contrast,
the $d$ of our non-rigid registration
(Figure~\ref{fig:quality_compare}\,b) has only a few spatial
frequencies with residual signal other than noise.

\subsection{Quantifying the Quality for Registration of a Series.}
Another potential metric for the quality of the registration would be
the absolute value of the average of $d$ in a small region.  Let
$p\in\mathbb{N}$. To evaluate how close $d$ is to noise, we average
$d$ over all patches of size $(2p+1)\times (2p+1)$ and consider the
absolute value of the average, resulting in an image $d_p$ of size
$(N-2p)\times(N-2p)$, \ie
\begin{equation}\label{eqn:d_p(index)}
d_p(\boldsymbol{i})=\abs{\frac{1}{(2p+1)^2}\sum_{\boldsymbol{k}\in P_p^{i_1+p,i_2+p}}d(\boldsymbol{k})}.
\end{equation}
Since the averaging over the patches reduces the noise in $d$, the
higher the quality of the registration and averaging procedure, the
smaller the values of $d_p$. Thus, the arithmetic mean of $d_p$ gives
a quantitative measure of the quality of the procedure, the lower the
better. In particular, this quantity allows for the comparison of the
quality of different registration
results. Figure~\ref{fig:quality_compare}d shows how $d_p$ for the
non-rigid registration is always smaller than $d_p$ for the manual
registration as p varies from 1 to 12.

To get a more local view of the quality measure, we can partition
$d_p$ in $L\times L$ patches of equal size (dropping pixels close to
the boundary if the width or height of $d_p$ is not a multiple of $L$)
and calculate the arithmetic mean in these patches instead of the
whole image. Figure~\ref{fig:quality_compare}e is a contour plot of
$d_2$ on a $9\times 9$ grid for the manual registration, while
Figure~\ref{fig:quality_compare}f is the same for the non-rigid
registration.  Note that the manual registration always performed more
poorly than the non-rigid registration even in the area used as the
reference for the manual registration.

The lower right hand portion of the reconstructions
(Figure~\ref{fig:high-res_compare}) correspond to an area of the
initial field of view where the sample was not in focus due to a
thickness gradient in the particular specimen.  Our algorithm
correctly registered this area of the frames on a coarse scale, but
did not artificially ``improve'' the image in this area.  The
reconstruction varies in quality across the field of view as the input
frames vary in information.  This is a distinct advantage over some
other image processing techniques which either enforce periodicity on
the images, or produce an ``average'' structure.


\subsection{Further Algorithm Extensions}
The above procedure should be viewed as one realization of a general concept that in principle allows one to extract additional information at the expense of additional computation. We briefly highlight two such possibilities. The first deals with single frame adjustments to the acquisition process. The second addresses an iterative extension of the above algorithm for a series of frames.

\subsubsection{A first order motion extraction.}
\label{sec:si:BasicDistortionModel}

The inevitable movement of the sample during the scanning process combined with how STEM rasters the electron probe on the sample to acquire an image leads to a spatial distortion in the STEM images.
In particular, Figure~\ref{fig:ninthDeformed}
shows a significant translation of the field of view encountered during the acquisition of the series of frames. In order to retain as much information as possible and ease the registration process by minimizing frame-to-frame distortions we introduce next a simple strategy for capturing the most prominent ``macroscale'' translational parts of the distortion.
To formulate this model we need to make assumptions on the movement of the sample and need to denote some quantities inherent to the STEM scanning process.

Let $t$ denote the time it takes to scan a single line in the image and let $t_f$ denote the flyback time, \ie the amount of time it takes to move the STEM probe from the end of one line to the beginning of the next line. As first order approximation, we assume that the sample undergoes a constant translational movement $v=(v_1,v_2)\in\mathbb{R}^2$ while a STEM image is acquired. The movement vector may differ from frame to frame though. Furthermore, we assume that the STEM probe starts to scan the sample at position $(0,0)$. Thus, when we scan the first pixel, corresponding to the position $(0,0)$, we also scan position $(0,0)$ of the sample. When scanning the last pixel in the first line, corresponding to the position $(1,0)$, the time $t$, \ie the time to scan a line, has passed and the sample has moved by $tv$. Hence, we are actually scanning the position $(1,0)-tv$ of the sample. Let $h$ denote the height of a STEM line as fraction of $1$, \ie $h=\frac{1}{N-1}$ where $N$ denotes the number of lines in the STEM image. Then, when scanning the first pixel of the second line, corresponding to the position $(0,h)$, the time $t+t_f$ has passed and we are actually scanning the position $(0,h)-(t+t_f)v$. Consequently, when scanning the last pixel of the second line, corresponding to the position $(1,h)$, the time $2t+t_f$ has passed and we are actually scanning the position $(1,h)-(2t+t_f)v$.

In general, when scanning a position $x_1\in[0,1]$ in line $k$, corresponding to $(x_1,kh)$, a time of $tx_1+(t+t_f)k$ has passed and we are actually scanning the position $(x_1,kh)-(tx_1+(t+t_f)k)v$. Therefore, assuming that the distortion is linearly interpolated between the lines, the image domain is deformed by the linear mapping
\[
x\mapsto
\begin{pmatrix}
1-tv_1&-\smallfrac{(t+t_f)v_1}{h}\\
-tv_2&1-\smallfrac{(t+t_f)v_2}{h}
\end{pmatrix}
x.
\]
In other words, denoting the matrix by $M$, instead of $f(x)$ the intensity acquired at position $x$ is actually showing $f(Mx)$. Thus, using the inverse of $M$, it would be possible to remove the distortion caused by a constant translation of the sample during the acquisition of the frame.

\subsubsection{An extended series averaging procedure.}
\label{subsec:extended_series}~
Algorithm~\ref{alg:SeriesAveraging} \notinclude{and its alterations described above} allows us to generate a reconstruction of the series that structurally closely resembles the first frame $f_1$ due to the fact that $f_1$ was used as initial reference frame in the algorithm. In particular, if we drop the outer loop, \ie for $K=1$, the reconstruction actually is a denoised version of $f_1$ (note that this is not guaranteed for $K\geq2$). Using a slight alteration of Algorithm~\ref{alg:SeriesAveraging}, we can register all frames to the $j$-th frame. This is achieved by applying the algorithm with $K=1$ and, without the averaging step, once on the series $f_{j-1},\ldots,f_1$ and once on the series $f_{j+1},\ldots,f_n$, each time using $f_j$ as reference frame. Having the transformations of all frames to the $j$-th frame, we can create a denoised reconstruction $\tilde{f}_j$ of the $j$-th frame and thus of every input frame. Furthermore, we can use the algorithm to register all of the denoised frames to the first of these frames $\tilde{f}_1$. Contrary to the initial registration of the input frames to $f_1$, the registration of the denoised frames can be done with a considerably lower regularization parameter $\lambda$ and at a higher precision since this registration is not hampered by the large amount of noise found in the input frames. Due to their improved accuracy, the deformations determined based on $\tilde{f}_1\ldots,\tilde{f}_n$ can be used to generate an improved reconstruction from our original input $f_1,\ldots,f_n$ by using these deformations when averaging the input frames with $f_1$ as reference frame. If desired, this process can even be iterated, \ie instead of only generating an improved reconstruction of $f_1$ one can generate an improved reconstruction of every frame and again create an improved average of the original data by registering the improved reconstructions and so on, which can be seen as an extension of the $k$-loop idea in Algorithm~\ref{alg:SeriesAveraging}.
Let us emphasize that all reconstructions here are always obtained by averaging the original noisy input frames, only the deformations used for the averaging were determined based on reconstructions we calculated previously.

The main drawback of this variant is the vastly increased computational cost compared to our original algorithm. Since the original algorithm essentially needs to be run $n+1$ times, once for each input frame to get a denoised version of this frame and once to create the improved reconstruction based on the registration of the denoised frames, the computational cost is increased by a factor of $n+1$. Thus, it is only feasible to use this approach if the computational cost is not a concern or if the number $n$ of input frames is moderate.

\section{Conclusions}\label{sec:Conclusion}
We have developed a strategy for extracting an increased level of information from a series of low dose STEM images,  rather than using single high dose images, in order to circumvent or at least significantly ameliorate a
build up of unwanted physical artifacts, contortions, and damage caused during the acquisition process. We have applied the methodology
to   beam sensitive materials like siliceous zeolite Y.

 A crucial ingredient  is
a non-linear registration process that removes visual inspection and human interaction.
The quantitatively improved automated information retrieval,
is an important step forward  since the huge amounts of data created by many modern imaging techniques employed in astrophysics, medical imaging, process control and various forms of microscopy call for an early and reliable triage before storage.
In particular, in many of these areas {\em change detection} is crucial and critically hinges on an accurate and reliable registration.

We provide algorithms and metrics that allow researchers to extract
larger amounts of meaningful information from data which often are
costly to acquire.  The quality of the final reconstruction hinges on
the quality of the initial input data.  In the case considered here of
a beam sensitive zeolite under low-dose conditions, significant improvement
in the information content of the reconstruction was demonstrated
relative to the individual input frames without artificially
``improving'' the areas in the field of view which either were not in
focus in the input frames or suffered significant beam damage during
the series acquisition.

\section{Acknowledgements}
This research was supported in part by the MURI/ARO Grant
W911NF-07-1-0185; the NSF Grants DMS-0915104 and DMS-1222390; the
Special Priority Program SPP 1324, funded by DFG; and National
Academies Keck Futures Initiative grant NAKFI IS11.





\bibliographystyle{model1-num-names}







\end{document}